\documentclass[review]{elsarticle}

\usepackage{times}
\usepackage{epsfig}
\usepackage{graphicx}
\usepackage{amsmath}
\usepackage{amssymb}
\usepackage{graphicx}
\usepackage{enumerate}
\usepackage{cases}
\usepackage{multirow}
\usepackage{verbatim}
\usepackage{amssymb}
\usepackage{CJK}
\usepackage{algorithm}
\usepackage{algorithmicx}
\usepackage{algpseudocode}

\usepackage{color}
\usepackage{bm}
\usepackage{booktabs}
\usepackage{hyperref}

\newcommand{\ie}{\textit{i}.\textit{e}.}
\newcommand{\eg}{\textit{e}.\textit{g}.}
\newcommand{\etc}{\textit{etc}.}

\journal{Neurocomputing}

\begin{document}

\begin{frontmatter}

\title{A Parallel Down-Up Fusion Network for Salient Object Detection in Optical Remote Sensing Images}
%
%

\author[add1]{Chongyi Li}
\ead{lichongyi@tju.edu.cn}

\author[add2,add3]{Runmin Cong\corref{mycorrespondingauthor}}
\cortext[mycorrespondingauthor]{Co-first and corresponding author}
\ead{rmcong@bjtu.edu.cn}

\author[add1]{Chunle Guo}
\ead{guochunle@tju.edu.cn}

\author[add4,add5]{Hua Li}
\ead{huali27-c@my.cityu.edu.hk}

\author[add2,add3]{Chunjie Zhang}
\ead{cjzhang@bjtu.edu.cn}

\author[add6]{Feng Zheng}
\ead{f.zheng@ieee.org}

\author[add2,add3]{Yao Zhao}
\ead{yzhao@bjtu.edu.cn}

\address[add1]{School of Electrical and Information Engineering, Tianjin University, Tianjin 300072, China}
\address[add2]{Institute of Information Science, Beijing Jiaotong University, Beijing 100044, China}
\address[add3]{Beijing Key Laboratory of Advanced Information Science and Network Technology, Beijing 100044, China}
\address[add4]{School of Software Engineering, Huazhong University of Science and Technology, Wuhan 430074, China}
\address[add5]{Department of Computer Science, City University of Hong Kong, Kowloon 999077, Hong Kong SAR, China}
\address[add6]{Department of Computer Science and Engineering, Southern University of Science and Technology, Shenzhen 518055, China}

\begin{abstract}
The diverse spatial resolutions, various object types, scales and orientations, and cluttered backgrounds in optical remote sensing images (RSIs) challenge the current salient object detection (SOD) approaches. It is commonly unsatisfactory to directly employ the SOD approaches designed for nature scene images (NSIs) to RSIs. In this paper, we propose a novel Parallel Down-up Fusion network (PDF-Net) for SOD in optical RSIs, which takes full advantage of the in-path low- and high-level features and cross-path multi-resolution features to distinguish diversely scaled salient objects and suppress the cluttered backgrounds. To be specific, keeping a key observation that the salient objects still are salient no matter the resolutions of images are in mind, the PDF-Net takes successive down-sampling to form five parallel paths and perceive scaled salient objects that are commonly existed in optical RSIs. Meanwhile, we adopt the dense connections to take advantage of both low- and high-level information in the same path and build up the relations of cross paths, which explicitly yield strong feature representations. At last, we fuse the multiple-resolution features in parallel paths to combine the benefits of the features with different resolutions, \ie, the high-resolution feature consisting of complete structure and clear details while the low-resolution features highlighting the scaled salient objects. Extensive experiments on the ORSSD dataset demonstrate that the proposed network is superior to the state-of-the-art approaches both qualitatively and quantitatively.
\end{abstract}


\end{frontmatter}


\section{Introduction}
\label{S:1}

\begin{figure}[!h]
  \centering
\begin{minipage}[b]{0.22\linewidth}
  \centering
  \centerline{\includegraphics[width=1\linewidth,height=1\linewidth]{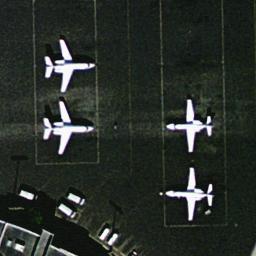}}
\end{minipage}
\begin{minipage}[b]{0.22\linewidth}
  \centering
  \centerline{\includegraphics[width=1\linewidth,height=1\linewidth]{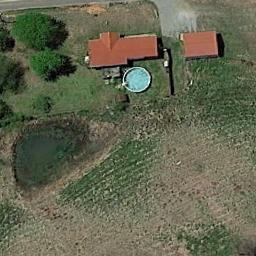}}
\end{minipage}
\begin{minipage}[b]{0.22\linewidth}
  \centering
  \centerline{\includegraphics[width=1\linewidth,height=1\linewidth]{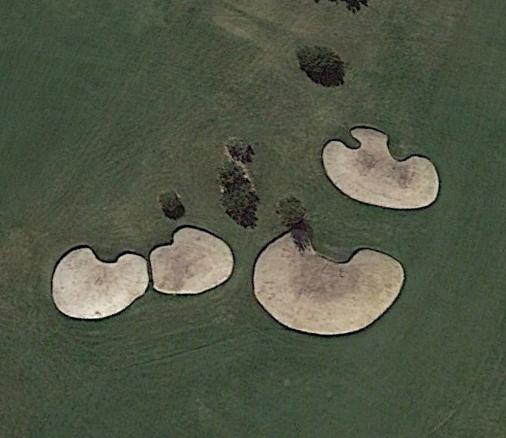}}
\end{minipage}
 \begin{minipage}[b]{0.22\linewidth}
  \centering
  \centerline{\includegraphics[width=1\linewidth,height=1\linewidth]{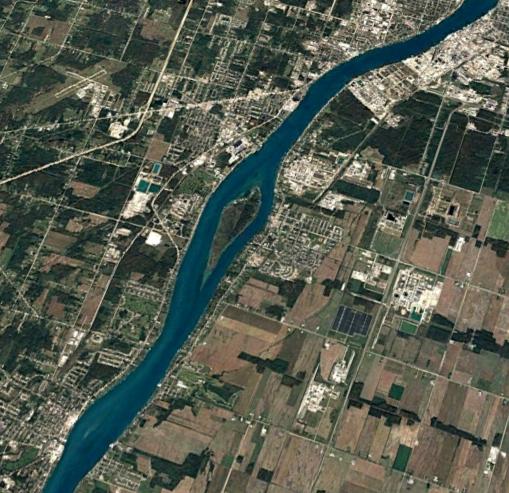}}
\end{minipage}
\caption{Several typical optical RSIs in human eye friendly color presentation.}
\label{fig:1}
\end{figure}

Human has the ability to automatically locate regions of interest from the complex scenes, which is called visual attention mechanism \cite{wwg_review}. Simulated by this mechanism, salient object detection (SOD) is to enable computers also to have a similar ability to automatically detect the most interesting and salient objects or regions in a scene. Due to its excellent scalability, SOD has been widely used in enhancement \cite{lcy2016tip,lcy2020tip}, foreground annotation \cite{R3}, segmentation \cite{R1,wwg_1}, retargeting \cite{R2,wwg_2}, quality assessment \cite{R5,jqp}, thumbnail creation \cite{R4}, and video summarization \cite{R6}. With different types of input data, SOD models can be classified into different categories, such as RGB SOD models \cite{R3Net,RFCN,PoolNet,aaai20}, RGB-D SOD models for RGB-D images \cite{rgbd1,rgbd2,rgbd3,lcy2020tc}, co-saliency detection models for image group \cite{icme18,cosal1,cosal2,cosal3}, video SOD for video sequences \cite{video1,video2,video3,fan2019Shifting}, light filed SOD models for light filed image \cite{lf1,lf2,lf3}, and remote sensing SOD models for remote sensing image (RSI) \cite{rs5,LVNet,hsi}. With the great need in military and civilian applications, the optical RSI has become more and more significant over the past few years. In spite of the rapid development of SOD for nature scene images (NSIs), less work focuses on SOD in optical RSIs, and thus, current SOD methods  are still far from being practical application in optical RSIs. The main reasons are that 1) the high-angle shot leads to the diversely scaled objects in optical RSIs, 2) the wide-scale devices lead to the various scenes and object types, and 3) the high-resolution optical RSIs contain cluttered backgrounds. The characteristics of optical RSIs are different from NSIs, which limits the direct application of existing SOD methods. Fig.~\ref{fig:1} shows some typical optical RSIs that are in human eye-friendly color presentation.

To solve the challenging issues of SOD in optical RSIs, we propose a parallel down-up fusion network, called PDF-Net. Considering the diversely scaled objects in optical RSIs, we take successive down-sampling to highlight the salient objects with different resolutions based on a key observation that the salient objects still are salient no matter the resolutions of images are. In addition, we build up the cross-path relations by passing the down-sampling features to the next path in a dense connections manner. With the multiple-resolution and saliency-related features in parallel paths, we up-sample these features to the same resolution and further integrate them for saliency prediction. Here, the sharp boundary, complete structure, and clear details in the high-resolution features and the highlighted salient objects in the low-resolution features are effectively combined. Besides, the in-path dense connections are used to combine the low-level detail information and high-level semantic information, which re-use the features and accelerate the gradient backpropagation. With the novel network architecture designed for SOD in optical RSIs, our method reaches the state-of-the-art performance and insights into the subsequent development of deep learning-based SOD in optical RSIs.

Compared with existing deep learning-based SOD approaches \cite{R3Net,RFCN,rgbd1,rgbd2}, our PDF-Net is specially designed for optical RSIs, where the above-mentioned characteristics of optical RSIs are fully considered. For example, we take successive down-sampling operations to highlight the diversely scaled salient objects in optical RSIs. For another example, we build up the relations of cross-path and in-path by dense connections, which diversifies the feature representations of our network, thus achieving good generalization capability for various scenes and object types.

To sum up, the main contributions are summarized as follows.

\begin{itemize}

   \item We propose a parallel down-up fusion network for SOD in optical RSIs. With this novel network architecture, the complementarity from cross-path and multi-resolution features can be sufficiently combined, which benefits for detecting the diversely scaled salient objects in optical RSIs.

   \item The dense connections  are introduced to SOD in optical RSIs, which integrate the high-level and low-level information and bridge the relations of cross paths, thus boosting the performance of SOD in optical RSIs.

   \item The proposed network can accurately and effectively locate the salient objects from the optical RSIs, even under challenging scenes such as cluttered backgrounds and multiple objects with diverse scales. Moreover, our method outperforms the state-of-the-art saliency detectors on the ORSSD dataset.

\end{itemize}

The rest of the paper is organized as follows. First, the related works on SOD are introduced in Section 2. Then, we provide the details of the proposed method in Section 3, mainly including the network architecture and loss function. The experimental comparisons and ablation analysis are conducted in Section 4. Finally, we conclude this paper in Section 6.

\section{Related Work} \label{sec2}

In this section, we first briefly introduce some representative SOD models designed for NSI, then review the SOD models specialized for optical RSI.

The past decades have witnessed the theoretical development and performance improvement of SOD for NSI \cite{wwg_review,video2,RERVIEW} and deep learning-based visual tasks \cite{lcy2018depth,wangcvpr20201,lowlight2020,wangcvpr20202,lcy2019tmm,xp2019,lcy2020pr,xp2020}. Initially, some hand-crafted features or visual priors, such as background prior \cite{RBD}, color contrast \cite{RC}, color compactness \cite{DCLC}, sparse representation \cite{DSR}, matrix decomposition \cite{SMD}, random walks \cite{RCRR}, \etc, are utilized to represent the saliency attribute of an object and generate the bottom-up SOD models. Recently, supervised by the labels, top-down SOD model is task-driven, including the popular deep learning based methods \cite{ R3Net, RFCN, PoolNet,aaai20,lcy2020tc,DCL, DSS, RADF,MMCI,TAN,fluid,RGBT1,RGBT2}.
Deng \emph{et al.} \cite{R3Net} designed some residual refinement blocks to recurrently achieve SOD.
Hu \emph{et al.} \cite{RADF} proposed a SOD network by exploiting the recurrently aggregated features from the FCN network.
Liu \emph{et al.} \cite{PoolNet} achieved a real-time SOD network by designing a feature pyramid network with two simple pooling-based modules.
Chen \emph{et al.} \cite{aaai20} considered the global context information and proposed a progressive aggregation network for SOD.
Li \emph{et al.} \cite{lcy2020tc} proposed an attention steered interweave fusion network for RGB-D salient object detection, where the depth information is used to improve the performance of RGB SOD.
Zhao \emph{et al.} \cite{fluid} proposed to employ the contrast prior to enhance the depth maps and integrate the RGB image and depth map by fluid pyramid for RGB-D SOD.
To take advantage of the complementary benefits fo RGB and thermal infrared images, Zhang \emph{et al.} \cite{RGBT1} proposed an end-to-end network for multi-modal SOD.

For the SOD in optical RSI, there is less relevant research.
Zhao \emph{et al.} \cite{rs5} achieved SOD in optical RSIs by using the sparse representation with the help of the global and background cues, and collected two optical RSI datasets with the corresponding pixel-level saliency masks, but both datasets are not publicly available.
Zhang \emph{et al.} \cite{rs8} proposed a feature fusion model under the low-rank matrix recovery framework to achieve SOD in RSIs.
For this task, Li \emph{et al.} \cite{LVNet} proposed the first deep learning-based method named LVNet, and released an optical remote sensing saliency detection (ORSSD) dataset with pixel-wise saliency ground truth. In this LVNet, the two-stream pyramid module learns the complementary features and local details, and the encoder-decoder module with nested connections determines the discriminative features to infer the saliency regions.
In addition, some related tasks in optical RSI also include the simple saliency unit, such as building extraction \cite{rs1}, Region-of-Interest (ROI) extraction \cite{rs2}, airport detection \cite{rs4}, oil tank detection \cite{rs6}, and ship detection \cite{rs7}.
Ma \emph{et al.} \cite{rs2} proposed a superpixel-to-pixel saliency model to assist in ROI extraction in remote sensing image.
Li \emph{et al.} \cite{rs1} designed some low-level features to find the salient regions and achieved building extraction method for remotely sensed images.
Zhang \emph{et al.} \cite{rs4} combined the vision- and knowledge-oriented saliencies to assist in determining the airport location in optical RSI.
Liu \emph{et al.} \cite{rs6} introduced a circular feature map to the low-level saliency model to achieve an unsupervised oil tank detection method.
Dong \emph{et al.} \cite{rs7} achieved the SOD by using the multi-scale and multi-orientation steerable pyramid, and proposed a multi-level ship detection method by integrating the region proposal generation and ship target identification.

\section{Proposed Method}
\label{sec3}

In this section, we first introduce the proposed PDF-Net architecture, consisting of a common feature extraction network, a down-up sampling with dense connections network, and a multi-resolution saliency feature fusion network. Then, the loss function that is used to drive the training of the proposed PDF-Net is given.

\begin{figure}[!t]
\centering
\centerline{\includegraphics[width=1\linewidth]{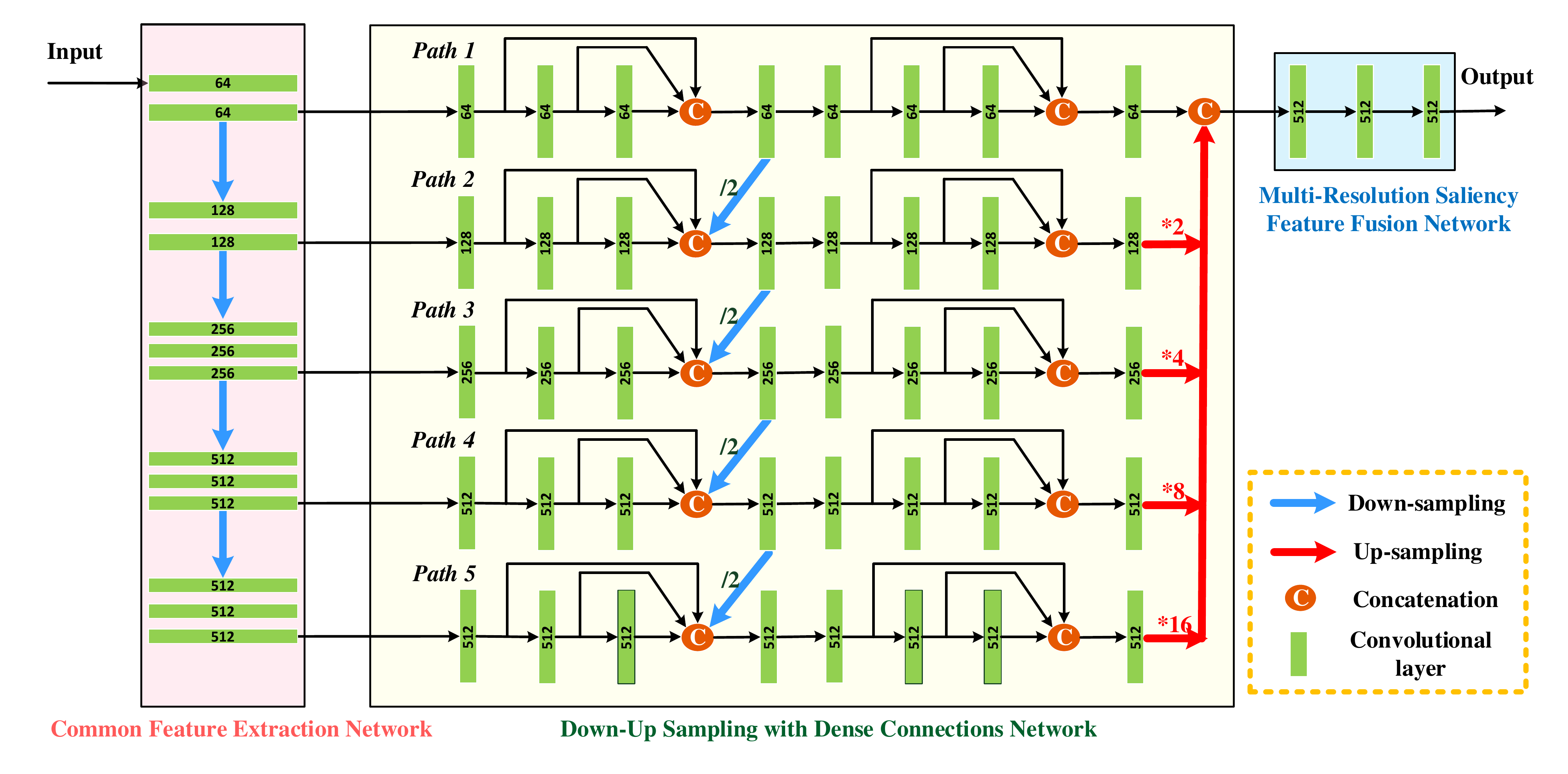}}
\caption{The overview framework of the proposed PDF-Net. The number of output features is indicated on the convolutional layers. The input is the optical RSI while the output represents the predicted saliency map. The number after the $/$ and $*$ indicates the times of down-sampling and up-sampling operations.  The down-sampling is implemented by the max-pooling layer with $2\times2$ filters and stride $2$. The up-sampling is implemented by linear interpolation. The concatenation indicates the features are concatenated along the channel dimension. All convolutional layers have  the kernels of size 3$\times$3 and stride 1}
\label{framework}
\end{figure}

\subsection{Overview}

The overview architecture of the proposed PDF-Net is presented in Fig.~\ref{framework}, which mainly includes a common feature extraction network, a down-up sampling with dense connections network, and a multi-resolution saliency feature fusion network. Specifically, an optical RSI is forwarded to the common feature extraction network that includes successive convolution and repeated down-sampling operations and then the common features are obtained for SOD. To address the challenging issue of diversely scaled salient objects in optical RSIs, we design a down-up sampling with dense connections network, where the extracted common features are passed through five parallel paths by the progressively down-sampling operation. In each parallel path, two dense connections units are adopted to further highlight the saliency-related features. With these highlighted features having different resolutions, we up-sample them to the same resolution as the input image and send them to the multi-resolution saliency feature fusion network. In the multi-resolution saliency feature fusion network, we further integrate the features by successive convolutions for accurate salient object prediction. As follows, we will introduce each network in detail.


\subsubsection{Common Feature Extraction Network}
For the SOD task, the VGG-16 \cite{VGG} or VGG-19 \cite{VGG} pre-trained on the ImageNet~\cite{ImageNet} was commonly employed as the common feature extraction network, which can increase the generalization capability of networks and also improve its accuracy. We use the VGG-16 as our common feature extraction network.  As shown in Fig.~\ref{framework}, the common feature extraction network includes four successive down-sampling operations and generates five groups of features with diverse resolutions. The common features before each down-sampling operation are forwarded to the down-up sampling with dense connections network, except for the last layer of the common feature extraction network where the features are directly forwarded to the fifth path of the down-up sampling with dense connections network.

\subsubsection{Down-up Sampling with Dense Connections Network}
In the optical RSIs, the type and scale of the salient objects vary diversely. Thus, we propose a down-up sampling with dense connections network, which is inspired by a key observation that the salient objects still are salient no matter the resolutions of an image are. To be specific, we form five parallel paths $p1$, $p2$, $p3$, $p4$, and $p5$, where the features have different resolutions. The path $p1$ keeps the same resolution as the input image.


In each path, the features are processed by two successive dense connections units, where each dense connection unit includes three successive convolution operations with the kernels of size 3$\times$3 and stride 1. Between two dense connections units in the same path, a convolutional layer (\ie, kernels size is 3$\times$3 and stride is set to 1) is used to compress the output features of the dense connections unit. Besides, the features between two dense connections units are down-sampled by using the max pooling layer with 2$\times$2 filters and stride 2. The down-sampled features are forwarded to the next path for building up the relations of different paths. For clearly understanding the operations in each path, we take path $p2$ as an example. Given the input features $F_{input}$, the first dense connections operation in  path $p2$ can be expressed as:
\begin{equation}
\label{equ_11}
F_{p2-1}=\sigma (\mathbf{W}_{p2-1}*F_{input}+\mathbf{b}_{p2-1}),
\end{equation}
\begin{equation}
\label{equ_12}
F_{p2-2}=\sigma (\mathbf{W}_{p2-2}*F_{p2-1}+\mathbf{b}_{p2-2}),
\end{equation}
\begin{equation}
\label{equ_13}
F_{p2-3}=\sigma (\mathbf{W}_{p2-3}*F_{p2-2}+\mathbf{b}_{p2-3}),
\end{equation}
\begin{equation}
\label{equ_14}
F_{con2-1}=Cat\{F_{p2-1},F_{p2-2},F_{p2-3},F_{p1-4}^{down}\},
\end{equation}
where $F_{p1-4}^{down}$ represents the 2$\times$ down-sampling features of the fourth convolutional layer in path $p1$, $Cat$ represents the concatenation operation along the channel dimension. After the first dense connections, the features $F_{con2-1}$ are fed to a convolutional layer for compressing the numbers of feature maps as:
\begin{equation}
\label{equ_15}
F_{p2-4}=\sigma (\mathbf{W}_{p2-4}*F_{con2-1}+\mathbf{b}_{p2-4}).
\end{equation}
Then, the features $F_{p2-4}$ are forwarded to the second dense connections unit as:
\begin{equation}
\label{equ_16}
F_{p2-5}=\sigma (\mathbf{W}_{p2-5}*F_{p2-4}+\mathbf{b}_{p2-5}),
\end{equation}
\begin{equation}
\label{equ_17}
F_{p2-6}=\sigma (\mathbf{W}_{p2-6}*F_{p2-5}+\mathbf{b}_{p2-6}),
\end{equation}
\begin{equation}
\label{equ_18}
F_{p2-7}=\sigma (\mathbf{W}_{p2-7}*F_{p2-6}+\mathbf{b}_{p2-7}),
\end{equation}
\begin{equation}
\label{equ_19}
F_{con2-2}=Cat\{F_{p2-5},F_{p2-6},F_{p2-7}\}.
\end{equation}
Next, the features $F_{con2-2}$  are fed to a convolutional layer for compressing the numbers of feature maps as:
\begin{equation}
\label{equ_20}
F_{p2-8}=\sigma (\mathbf{W}_{p2-8}*F_{con2-2}+\mathbf{b}_{p2-8}),
\end{equation}
At last, we up-sample the features $F_{p2-8}$ by a scale of 2.
The number of output features in each convolutional layer is indicated in Fig.~\ref{framework}.

In our down-up sampling with dense connections network, we finally up-sample the features at the end of each path to the same resolution as the input image and concatenate them, which can be represented as:
\begin{equation}
\label{equ_14}
F_{fusion}=Cat\{F_{path1},F_{path2},F_{path3},F_{path4},F_{path5}\},
\end{equation}
where $F_{path1}$, $F_{path2}$, $F_{path3}$, $F_{path4}$, and  $F_{path5}$ represent the features at the end of path 1, 2, 3, 4, and 5, respectively.  The purpose of fusing the features from parallel paths is to combine the complementary features. Usually, the high-resolution features contain complete and clear details while the low-resolution features can highlight the salient objects well. The combination of different resolution features can advantage in addressing the scale variation and perceiving the local details of the salient objects in optical RSIs.

\subsubsection{Multi-resolution Saliency Feature Fusion Network}
With the multi-resolution fusion features, we further suppress the backgrounds and highlight salient objects in the multi-resolution saliency feature fusion network, which can be described as:
\begin{equation}
\label{equ_15}
F_{m1}=\sigma (\mathbf{W}_{m1}*F_{fusion}+\mathbf{b}_{m1}),
\end{equation}
\begin{equation}
\label{equ_16}
F_{m2}=\sigma (\mathbf{W}_{m2}*F_{m1}+\mathbf{b}_{m2}),
\end{equation}
\begin{equation}
\label{equ_17}
Output=\delta (\mathbf{W}_{m3}*F_{m2}+\mathbf{b}_{m3}),
\end{equation}
where $\delta$ represents the Sigmoid activation function for predicting saliency map. Here, similar with the common feature extraction network, each convolutional layer with kernel of size 3$\times$3 and stride 1 followed by the ReLU  {\color{red}activation} function outputs 512-dimension features, except for the last layer that is followed by the Sigmoid {\color{red}activation} function and outputs the saliency map.

\subsection{Loss Function}
We follow the standard cross-entropy loss to optimize the proposed PDF-Net, which is formulated as:
\begin{equation}
\label{equ_1}
Loss=-(ylog(z)+(1-y)log(1-z)),
\end{equation}
where $y$ denotes the ground truth, and $z$ represents the predicted result.

\section{Experiments}\label{sec4}

In this section, we first introduce the benchmark dataset, evaluation metrics, training strategies, and implementation details. Then, we carry out some experiments to demonstrate the effectiveness of the proposed method, including performance comparisons and ablation studies.

\subsection{Dataset and Evaluation Metrics}

To verify the effectiveness and advantages of the proposed method, we train and test our method on the only publicly available ORSSD dataset\footnote{\url{https://li-chongyi.github.io/proj_optical_saliency.html}} \cite{LVNet} for SOD in optical RSIs. The ORSSD dataset includes 600 training images and 200 testing images, which is extremely challenging due to its diverse spatial resolution, the cluttered background, diverse object types, and the variable number and size of salient objects.


For the quantitative evaluations, we introduce the Precision-Recall (P-R) curve, F-measure, MAE score, and S-measure in experiments.
The P-R curve is drawn based on different combination of precision and recall scores, where the precision and recall scores are calculated by comparing the binary saliency map under different segmentation thresholds with ground truth label. The closer the P-R curve is to (1,1), the better the performance of the method. In addition, we can integrate the precision and recall scores as a comprehensive measurement by the weighted harmonic averaging, which is defined as F-measure \cite{Fmeasure2}:
\begin{equation}
\label{F}
F_{\beta}=\frac{(1+\beta^{2})Pre\times Rec}{\beta^{2}\times Pre+ Rec},
\end{equation}
where $Pre$ and $Rec$ correspond to the precision score and recall score, respectively, and $\beta^{2}$ is set to $0.3$ for emphasizing the precision as suggested in \cite{RERVIEW}. With a larger F-measure value indicating a better performance.

If we directly compare the saliency map $S$ and ground truth $GT$, the calculated difference is defined as MAE score \cite{MAE}:
\begin{equation}
\label{MAE}
MAE =\frac{1}{W\times H} \sum_{m=1}^{W} \sum_{n=1}^{H} |S(m,n)-GT(m,n)|,
\end{equation}
where $H$ and $W$ represent the height and width of the input image, respectively, and a smaller value indicates a smaller gap hence better.

In order to describe the structural attribute of the result, S-measure \cite{S-measure} is designed by combining the region similarity and object similarity as:
\begin{equation}
\label{Sm}
S_m = \alpha \times S_o+(1-\alpha) \times S_r,
\end{equation}
where $S_r$ and $S_o$ correspond to the region similarity and object similarity, respectively, $\alpha$ is set to $0.5$ as suggested in \cite{S-measure}, and a larger $S_m$ means the better performance.

\subsection{Training Strategies and Implementation Details}

Following the \cite{LVNet}, we use the same training images in ORSSD dataset to train our model, and test on the testing subset. In our implementations, the training samples are augmented by flipping and rotation and all the samples are resized to a fixed size of $128\times 128$ due to our limited memory. Our PDF-Net is implemented by TensorFlow on a PC with an Nvida GTX $1080$Ti GPU. During the training procedure, the batch size is set to $8$, the filter weights of each layer are initialized by standard Gaussian distribution, the bias is initialized as constant. We use ADAM \cite{Kingma2014} for network optimization and fix the learning rate to $1e^{-4}$.

\subsection{Comparison with State-of-the-art Methods}
\begin{figure}[!t]
\centering
\centerline{\includegraphics[width=1\linewidth]{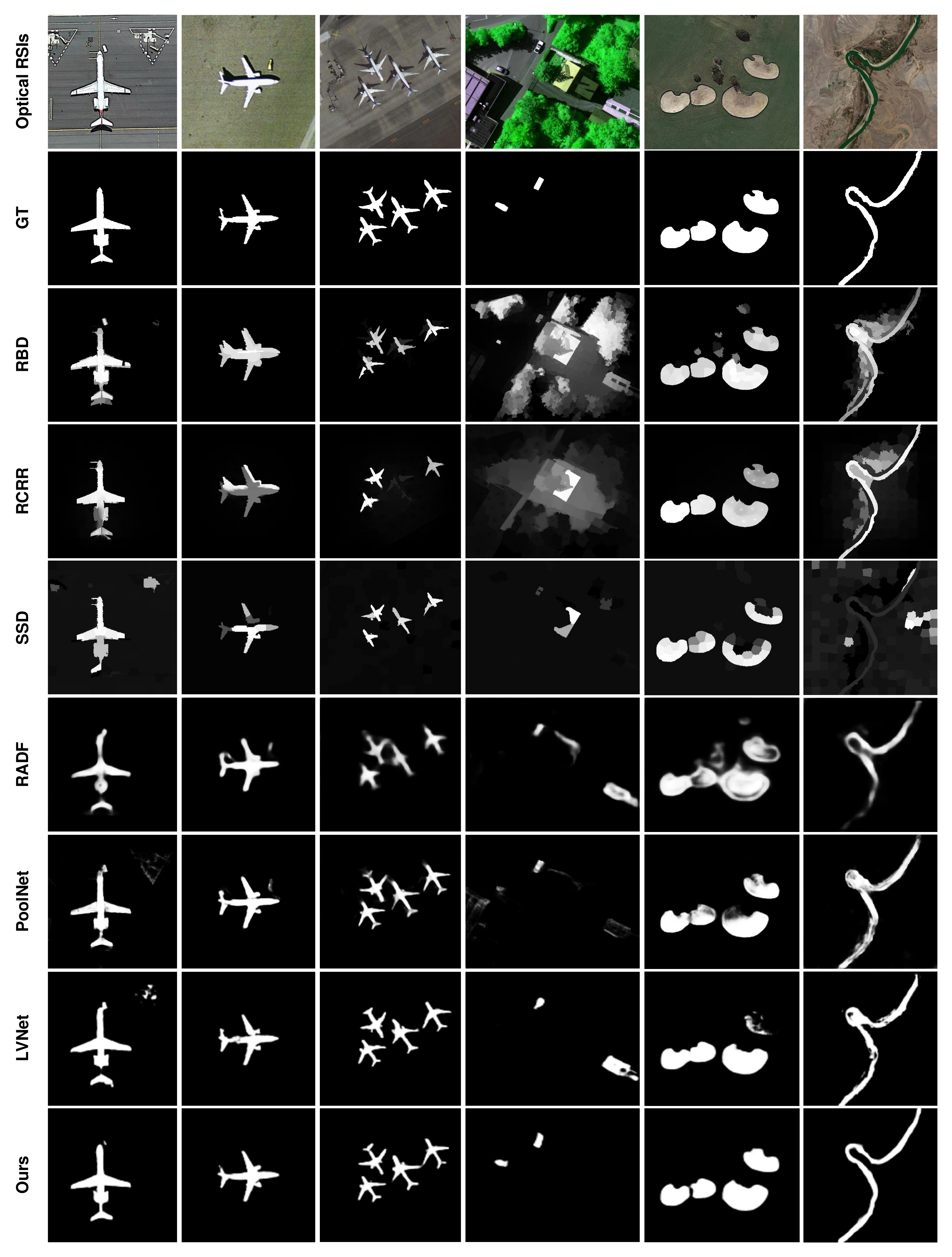}}
\caption{Visual examples of different methods. From top to bottom are the input optical RSIs, the corresponding ground truth images, the results of RBD \cite{RBD}, RCRR \cite{RCRR}, SSD \cite{rs5}, RADF \cite{RADF}, PoolNet \cite{PoolNet}, LVNet \cite{LVNet} and ours.}
\label{visual}
\end{figure}

In the experiments, we compare the proposed PDF-Net with 13 state-of-the-art SOD methods on the testing subset of the ORSSD dataset, including four unsupervised methods for NSIs (\ie, RBD \cite{RBD}, DSG \cite{DSG}, MILPS \cite{MILPS}, and RCRR \cite{RCRR}), five deep learning-based methods for NSIs (\ie, DSS \cite{DSS}, RADF \cite{RADF}, R3Net \cite{R3Net}, RFCN \cite{RFCN}, and PoolNet \cite{PoolNet}), and four methods for optical RSIs (\ie, SSD \cite{rs5}, SPS \cite{rs2}, ASD \cite{rs4}, and LVNet \cite{LVNet}). For a fair comparison, the results of competitors are generated by the released codes or directly provided by the authors.Moreover, the deep learning-based methods for NSIs are retrained on the same training data of the ORSSD dataset using their default parameter settings.

Some visual comparisons of different methods are shown in Fig. \ref{visual}, including seven examples with different salient objects (\eg, ship, river, buildings, stones, and airplanes). From the visual comparisons, our proposed method shows advantages in the following aspects.

\begin{enumerate}[(1)]

   \item \textbf{Our method can detect salient objects more accurately and completely, and suppress the background regions more cleanly and effectively.} In the first image, the proposed method can completely detect the aircraft, while suppressing the background region well. However, other methods have drawbacks in these aspects, such as the SSD method \cite{rs5} fails to suppress the background regions in the top right, the deep learning-based methods (\eg, LVNet \cite{LVNet} and PoolNet \cite{PoolNet}) also include some background noises in these regions. In addition, the structure of the aircraft detected by some methods is not complete, such as SSD \cite{rs5} and RADF \cite{RADF}. Similarly, in the second image, the proposed method can detect salient objects more meticulously and completely. For example, compared with the LVNet \cite{LVNet} method, the structure of the aircraft is more complete and sharper (\eg, the tail) in our result.

   \item \textbf{Our method performs well in small and multiple objects scenarios.} In the fourth image, the salient ship objects are extremely small, thus it is very challenging for SOD methods. For the unsupervised methods (\eg, RBD \cite{RBD}, RCRR \cite{RCRR}, and SSD \cite{rs5}), they are completely incapable of effectively detecting the salient objects. The deep learning-based methods (\eg, RADF \cite{RADF} PoolNet \cite{PoolNet}, and LVNet \cite{LVNet}) can detect one of the two cars, but the background regions have limited capability of suppression and include more background noises (\eg, the building at the bottom right). By contrast, our method can basically locate these two cars, and the backgrounds are effectively suppressed. In the third image, the RCRR \cite{RCRR} misses one aircraft, and some methods (\eg, RBD \cite{RBD}, SSD \cite{rs5}, and RADF \cite{RADF}) basically could not completely detect the entire aircraft. By contrast, our method obtains more complete and accurate detection result. In the fifth image,  RBD \cite{RBD} method fails to suppress the background effectively, and some methods (\eg, SSD \cite{rs5}, RADF \cite{RADF}, PoolNet \cite{PoolNet}, and LVNet \cite{LVNet}) can not detect the salient objects completely and exhaustively. Interestingly, the unsupervised RCRR method \cite{RCRR} performs better in this scenario. However, our method is better than the RCRR method \cite{RCRR} in terms of the internal consistency of salient objects.

   \item \textbf{Our method performs well in cluttered and complex scenes.} In the last image, the background is complex, which causes most methods not accurately and completely detecting the salient region, and many background regions are incorrectly retained. By contrast, our method owns better detection performance with clean backgrounds, fine boundaries, and complete foregrounds.

\end{enumerate}

 In summary, the proposed PDF-Net achieves  better visual performance in terms of accurate location, complete structure, sharp boundary, and clear background. This mainly benefits from the specialized design of our PDF-Net for SOD in optical RSIs, especially the down-up sampling with dense connections network.\par

\begin{figure}[!t]
\centering
\centerline{\includegraphics[width=0.8\linewidth]{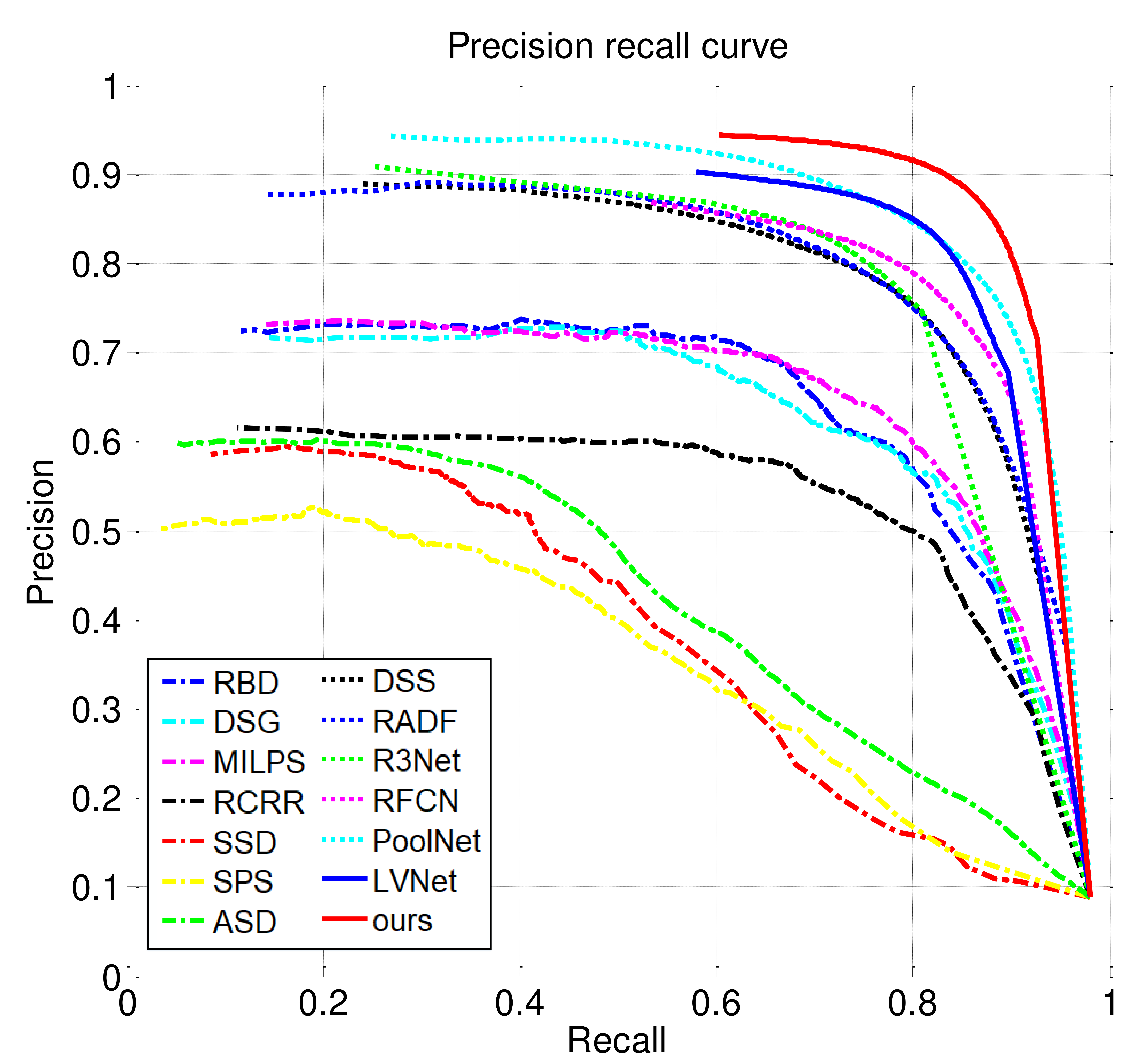}}
\caption{Illustration of P-R curves on the test subset of ORSSD dataset. }
\label{PR}
\end{figure}

\begin{table}[!t]
\renewcommand\arraystretch{1}
\caption{Quantitative evaluations of different methods on the testing subset of ORSSD dataset. The best and second best performances are marked in red and blue, respectively.}
\begin{center}
\setlength{\tabcolsep}{4mm}{
\begin{tabular}{c|c|c|c|c|c}
\hline
Method & Precision & Recall & $F_{\beta}$ & MAE & $S_m$ \\
\hline\hline

RBD \cite{RBD} & $0.7080$ & $0.6268$ & $0.6874$ & $0.0626$ & $0.7662$ \\
\hline
RCRR \cite{RCRR} & $0.5782$ & $0.6552$ & $0.5944$ & $0.1277$ & $0.6849$ \\
\hline
DSG \cite{DSG} & $0.6843$ & $0.6007$ & $0.6630$ & $0.1041$ & $0.7195$ \\
\hline
MILPS \cite{MILPS} & $0.6954$ & $0.6549$ & $0.6856$ & $0.0913$ & $0.7361$ \\
\hline
SSD \cite{rs5} & $0.5188$ & $0.4066$ & $0.4878$ & $0.1126$ & $0.5838$ \\
\hline
SPS \cite{rs2} & $0.4539$ & $0.4154$ & $0.4444$ & $0.1232$ & $0.5758$ \\
\hline
ASD \cite{rs4} & $0.5582$ & $0.4049$ & $0.5133$ & $0.2119$ & $0.5477$ \\
\hline
R3Net \cite{R3Net} & $0.8386$ & $0.6932$ & $0.7998$ & $0.0399$ & $0.8141$ \\
\hline
DSS \cite{DSS} & $0.8125$ & $0.7014$ & $0.7838$ & $0.0363$ & $0.8262$ \\
\hline
RADF \cite{RADF} & $0.8311$ & $0.6724$ & $0.7881$ & $0.0382$ & $0.8259$  \\
\hline
RFCN \cite{RFCN} & $0.8239$ & $0.7376$ & $0.8023$ & $0.0293$ & $0.8437$  \\
\hline
PoolNet \cite{PoolNet} & $\textcolor[rgb]{0.00,0.07,1.00}{0.8799}$ & $0.7363$ & $\textcolor[rgb]{0.00,0.07,1.00}{0.8420}$ & $0.0293$ & $0.8551$  \\
\hline
LVNet \cite{LVNet} & $0.8672$ & $\textcolor[rgb]{0.00,0.07,1.00}{0.7653}$ & $0.8414$ & $\textcolor[rgb]{0.00,0.07,1.00}{0.0207}$ & $\textcolor[rgb]{0.00,0.07,1.00}{0.8815}$ \\
\hline
Ours & $\textcolor[rgb]{1.00,0.00,0.00}{0.9144}$ & $\textcolor[rgb]{1.00,0.00,0.00}{0.8027}$ & $\textcolor[rgb]{1.00,0.00,0.00}{0.8860}$ & $\textcolor[rgb]{1.00,0.00,0.00}{0.0149}$ & $\textcolor[rgb]{1.00,0.00,0.00}{0.9112}$ \\
\hline
\end{tabular}}
\end{center}
\label{tab1}
\end{table}

The P-R curves of different methods are shown in Fig. \ref{PR}. Observing it, we can see that the proposed PDF-Net achieves a higher position compared with the other 13 methods, followed by PoolNet and LVNet. Moreover, the deep learning-based methods are generally superior to the unsupervised methods. In order to more intuitively compare the performance of different methods, we report the numerical comparisons in Table \ref{tab1}, mainly including the Precision score, Recall score, F-measure, MAE score, and S-measure. For the unsupervised method, RBD method \cite{RBD} exhibits competitive performance, and its F-measure reaches 0.6874. It should be noted that the F-measure of the method specifically designed for optical RSIs is only 0.4878, which is very unsatisfactory. For the 5 deep learning based methods for NSIs with optical RSIs retraining, the F-measure can be improved to close to 0.8, and even the latest PoolNet method \cite{PoolNet} achieves the sub-optimal performance in all comparison methods. In contrast, the specially designed saliency networks for optical RSI (\ie, LVNet \cite{LVNet}) captures the second best performance in terms of Recall score, MAE score, and S-measure, which also illustrates the importance of the further development of SOD in optical RSI. From Table \ref{tab1}, we can see that our method achieves the best quantitative performance in all measurements. Specifically, compared with the \emph{\textbf{second best method}}, the percentage gain of our proposed method reaches $5.2\%$ for F-measure, $28.0\%$ for MAE score, and $3.4\%$ for S-measure. Besides, compared with the latest SOD in optical RSIs model LVNet \cite{LVNet}, our PDF-Net (228.32M) has the comparable model sizes with the LVNet (221.16M). All these quantitative evaluations demonstrate the superiority and effectiveness of the proposed method.

\subsection{Ablation Analysis}

To demonstrate the effectiveness obtained by the key components in the proposed PDF-Net, we carry out four ablation studies.

\begin{itemize}
\item  PDF-Net without the dense connections (\textbf{PDF-Net w/o DC}), \ie, the concatenation operations along the channel dimension in each path are all removed and thus the features are directly forwarded to the next convolutional layer.
\item PDF-Net only with one dense connections unit (including three successive convolutional layers and one concatenation operations) in each path (\textbf{PDF-Net w 1-DC}), \ie, the first dense connections unit in each path is removed.
\item PDF-Net without the cross-path connections (\textbf{PDF-Net w/o CPC}), \ie, the down-sampling features between two dense connections units are no more passed to the next path, and thus, cut off the relations between the adjacent paths.
\item We separately extract the features with original resolutions in five parallel paths (\textbf{PDF-Net w/o DUS}), \ie, the down-sampling and up-sampling operations are removed and all operations are conducted on the features of original resolution in each path.
\end{itemize}

For a fair comparison, we follow the identical network parameters as the PDF-Net, except for the modified parts.  The quantitative results for the ablation analysis on the testing subset of the ORSSD dataset are presented in Table \ref{tab2}.

\begin{table}[!t]
\renewcommand\arraystretch{1}
\caption{Quantitative results of ablation studies on the testing subset of ORSSD dataset.}
\begin{center}
\setlength{\tabcolsep}{3mm}{
\begin{tabular}{c|c|c|c}
\hline
  & $F_{\beta}$ & MAE & $S_m$\\
\hline\hline
PDF-Net & 0.8860& 0.0149&0.9112 \\
\hline
PDF-Net w/o DC &0.7980 &0.0381 &0.8367 \\
\hline
PDF-Net w 1-DC &0.8197 &0.0352 &0.8378 \\
\hline
PDF-Net w/o CPC &0.8466 &0.0244 &0.8704 \\
\hline
PDF-Net w/o DUS &0.8519 &0.0242 &0.8724 \\
\hline
\end{tabular}}
\end{center}
\label{tab2}
\end{table}

\begin{figure}[!t]
  \centering
\begin{minipage}[b]{0.13\linewidth}
  \centering
  \centerline{\includegraphics[width=1\linewidth,height=0.9\linewidth]{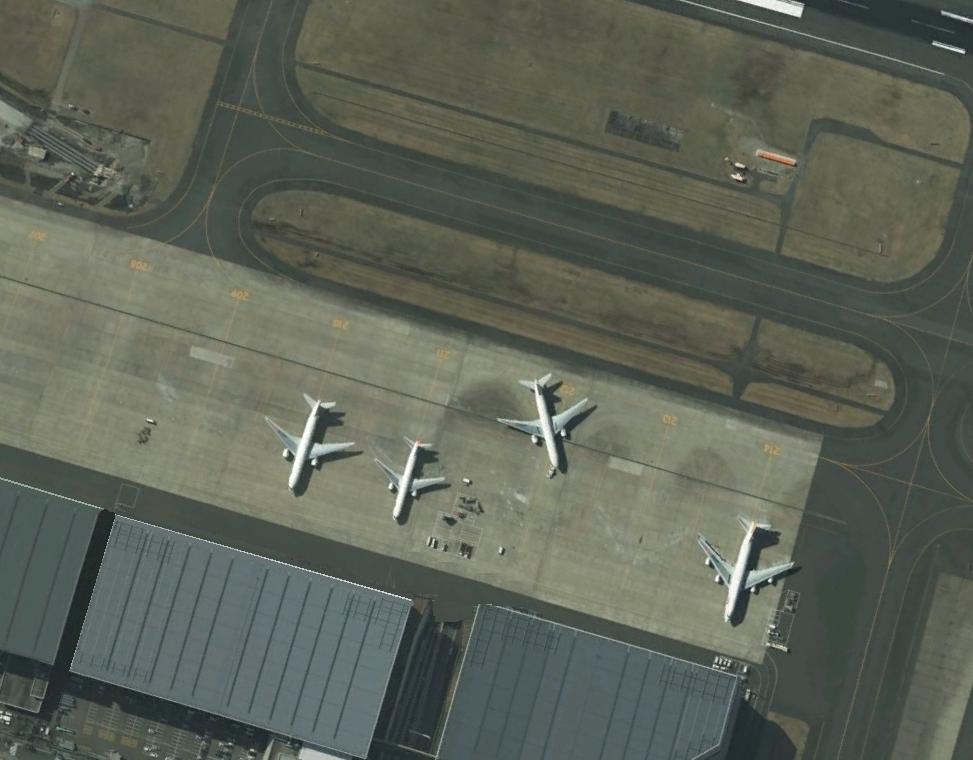}}
\end{minipage}
\begin{minipage}[b]{0.13\linewidth}
  \centering
  \centerline{\includegraphics[width=1\linewidth,height=0.9\linewidth]{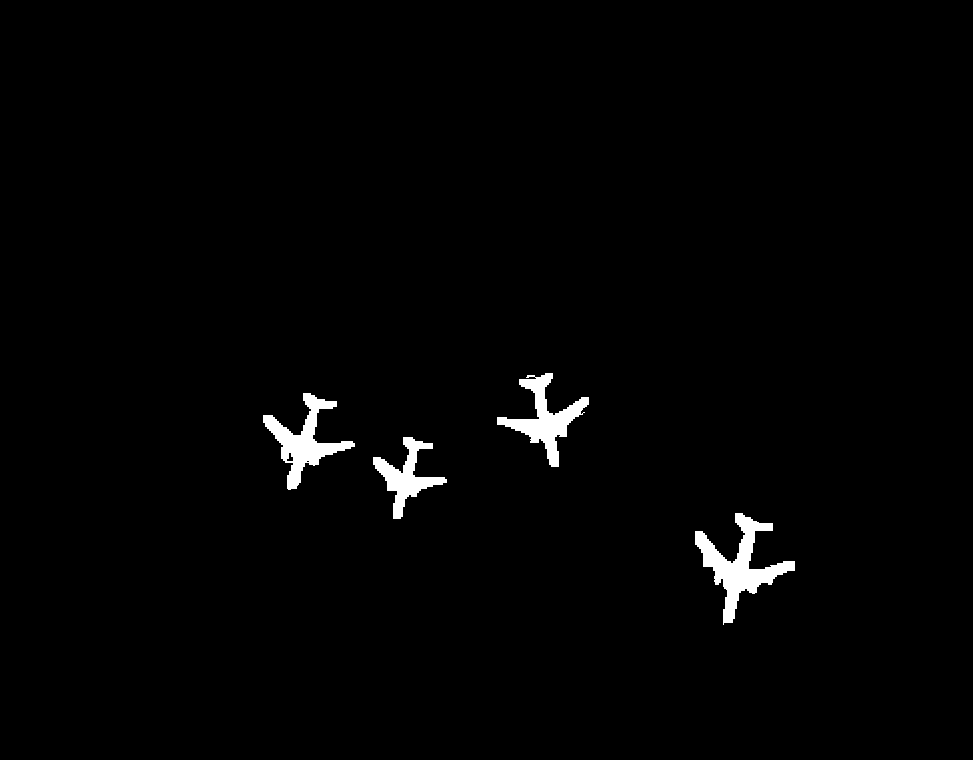}}
\end{minipage}
 \begin{minipage}[b]{0.13\linewidth}
  \centering
  \centerline{\includegraphics[width=1\linewidth,height=0.9\linewidth]{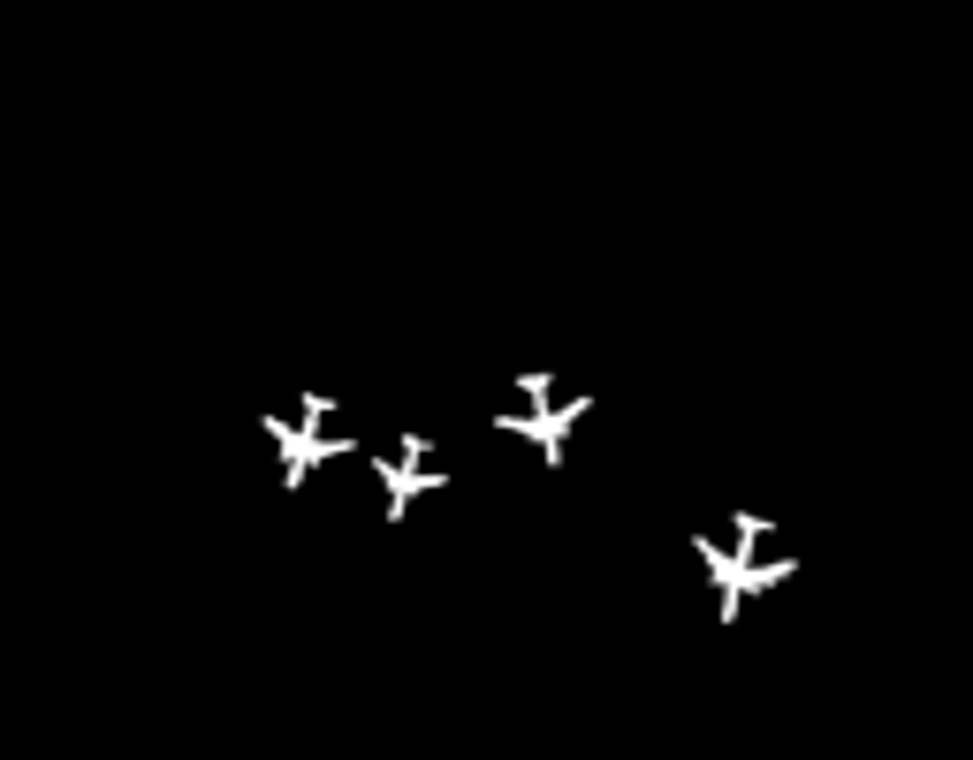}}
\end{minipage}
 \begin{minipage}[b]{0.13\linewidth}
  \centering
  \centerline{\includegraphics[width=1\linewidth,height=0.9\linewidth]{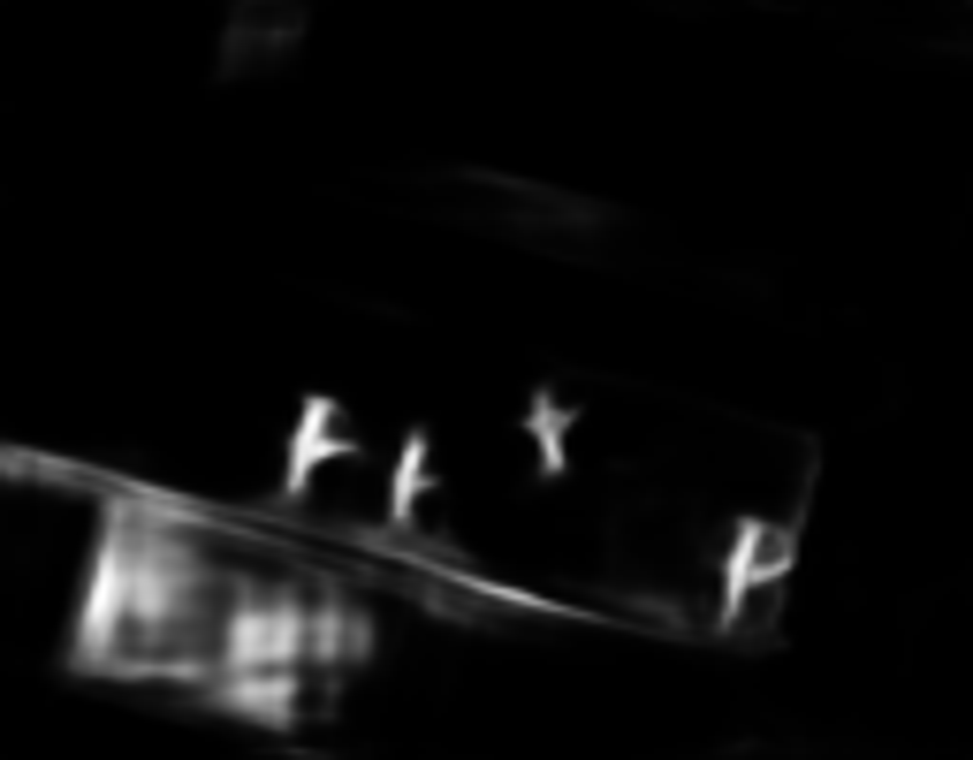}}
\end{minipage}
 \begin{minipage}[b]{0.13\linewidth}
  \centering
  \centerline{\includegraphics[width=1\linewidth,height=0.9\linewidth]{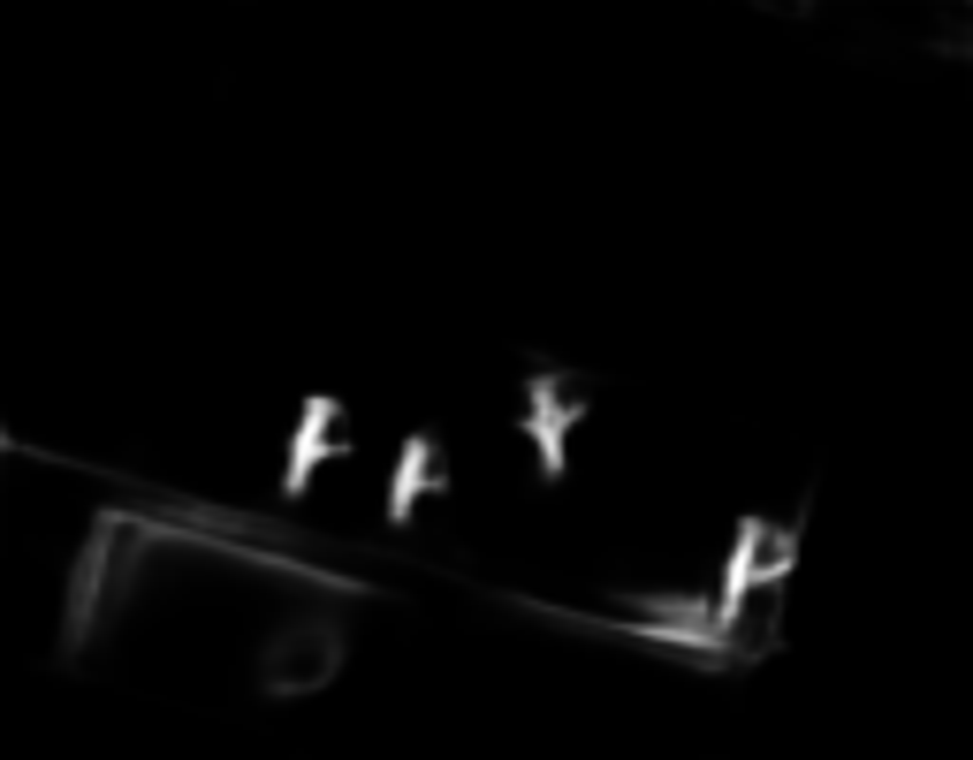}}
\end{minipage}
 \begin{minipage}[b]{0.13\linewidth}
  \centering
  \centerline{\includegraphics[width=1\linewidth,height=0.9\linewidth]{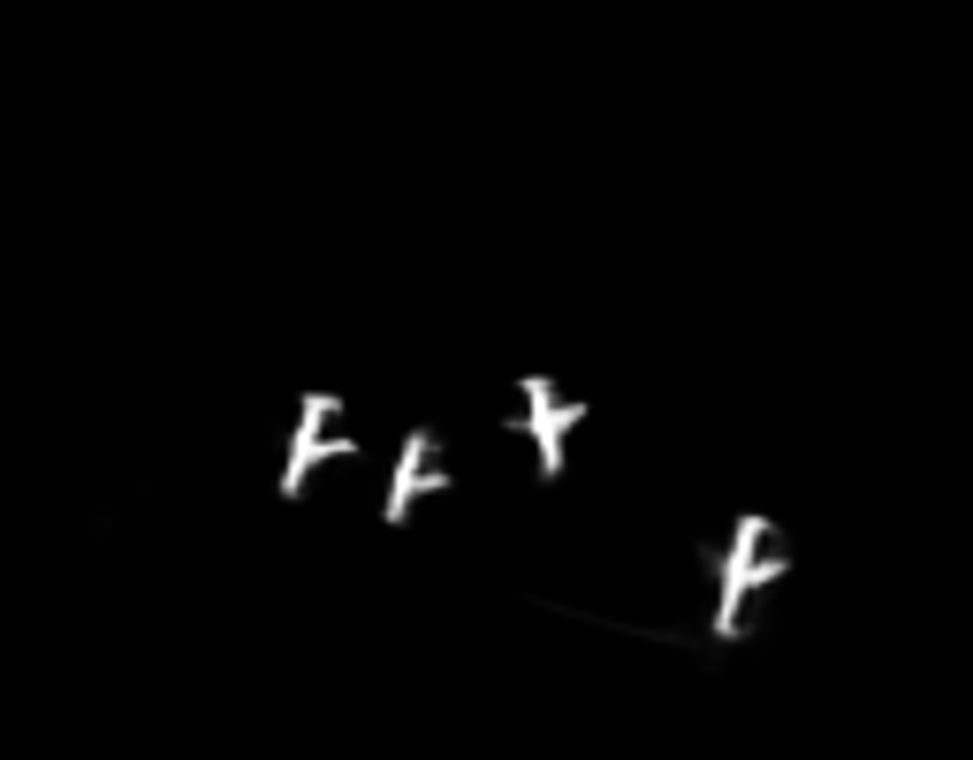}}
\end{minipage}
 \begin{minipage}[b]{0.13\linewidth}
  \centering
  \centerline{\includegraphics[width=1\linewidth,height=0.9\linewidth]{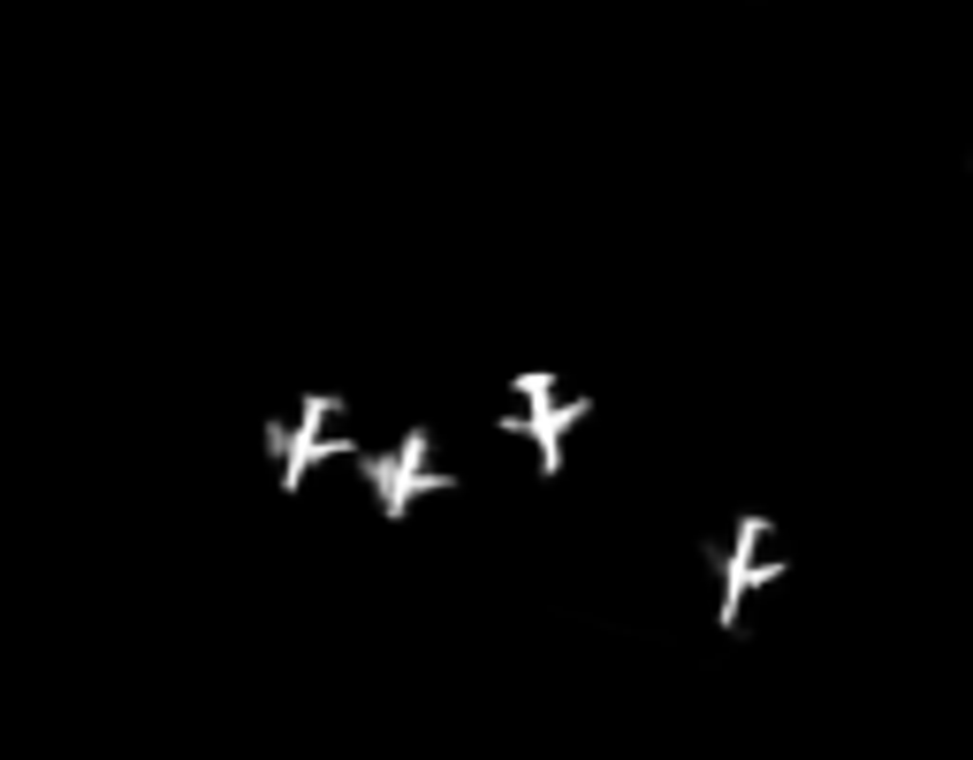}}
\end{minipage} \\

\begin{minipage}[b]{0.13\linewidth}
  \centering
  \centerline{\includegraphics[width=1\linewidth,height=0.9\linewidth]{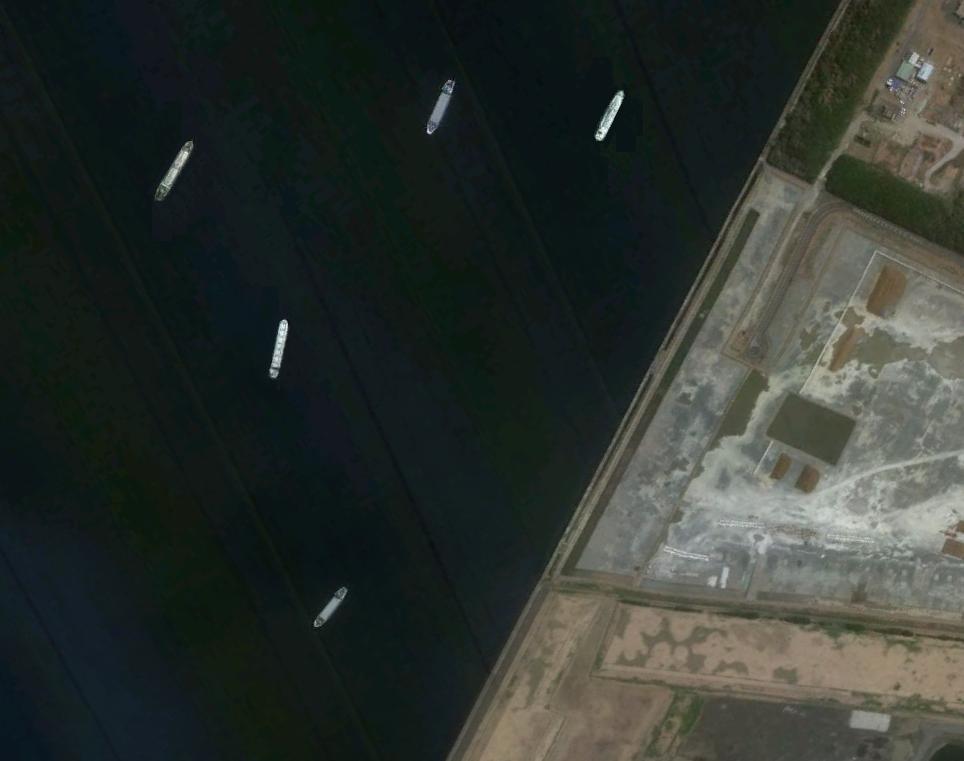}}
\end{minipage}
\begin{minipage}[b]{0.13\linewidth}
  \centering
  \centerline{\includegraphics[width=1\linewidth,height=0.9\linewidth]{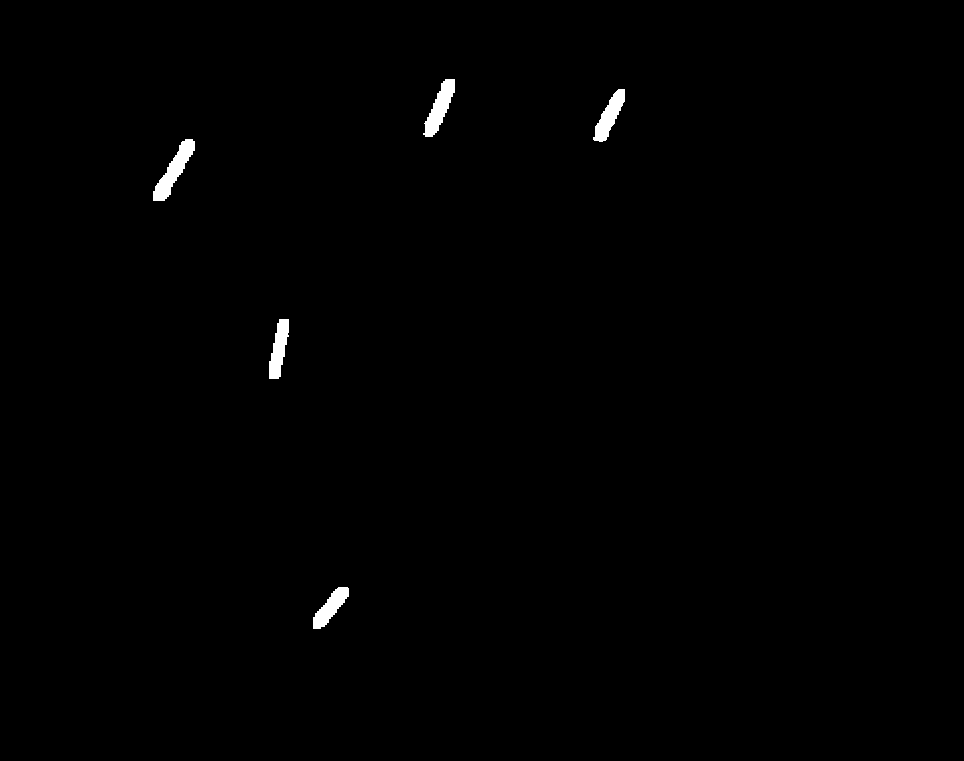}}
\end{minipage}
 \begin{minipage}[b]{0.13\linewidth}
  \centering
  \centerline{\includegraphics[width=1\linewidth,height=0.9\linewidth]{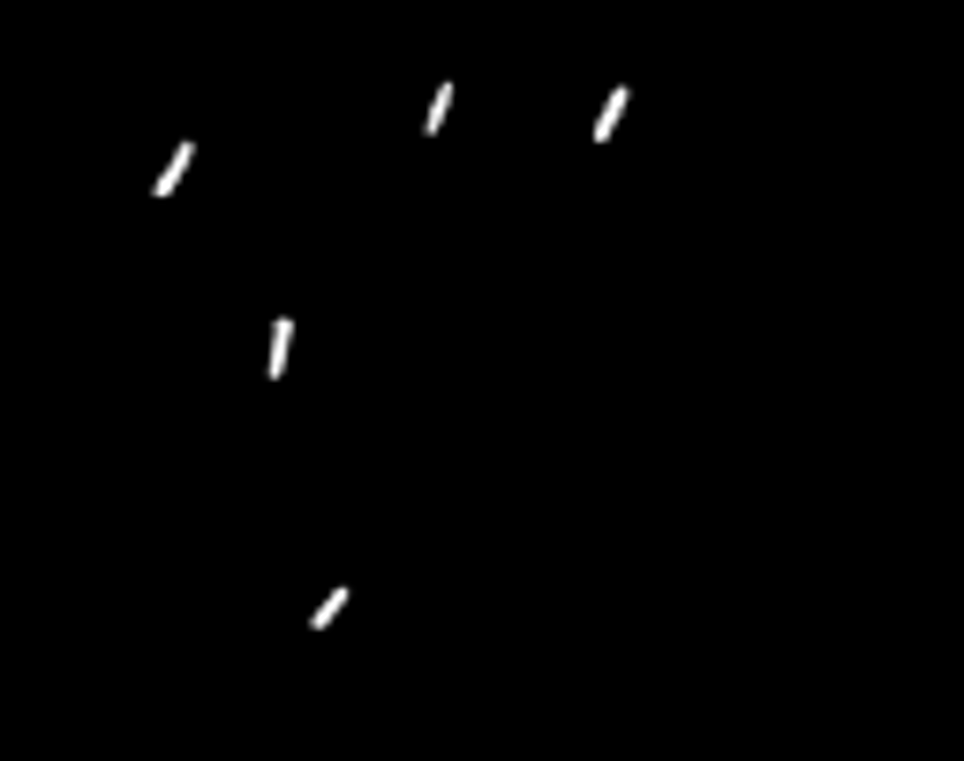}}
\end{minipage}
 \begin{minipage}[b]{0.13\linewidth}
  \centering
  \centerline{\includegraphics[width=1\linewidth,height=0.9\linewidth]{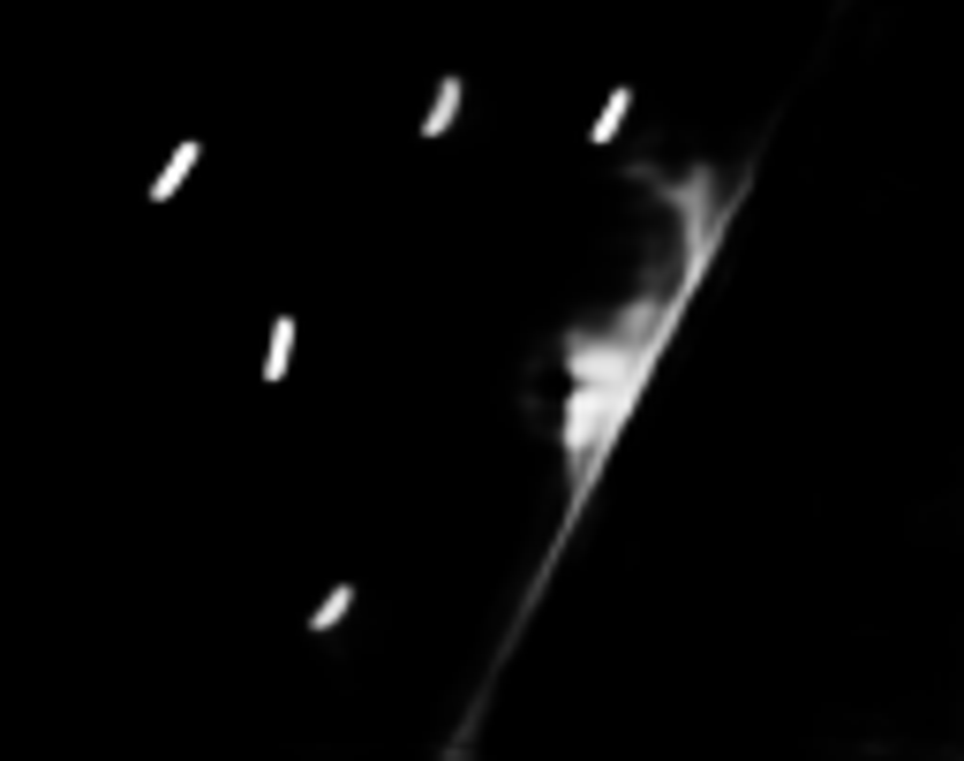}}
\end{minipage}
 \begin{minipage}[b]{0.13\linewidth}
  \centering
  \centerline{\includegraphics[width=1\linewidth,height=0.9\linewidth]{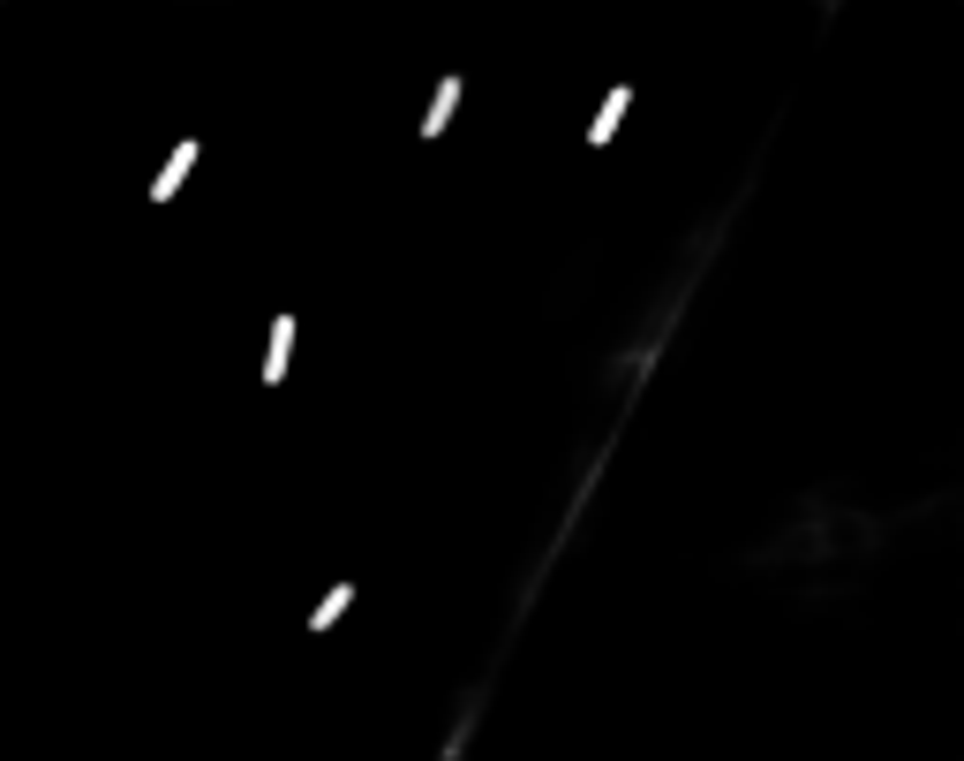}}
\end{minipage}
 \begin{minipage}[b]{0.13\linewidth}
  \centering
  \centerline{\includegraphics[width=1\linewidth,height=0.9\linewidth]{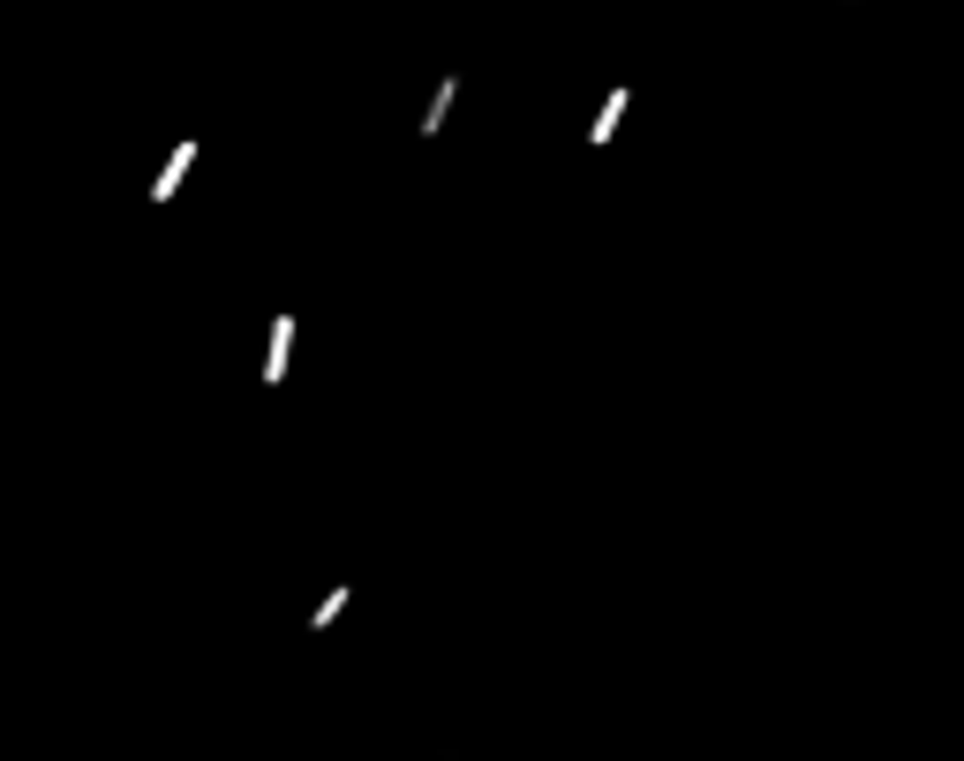}}
\end{minipage}
 \begin{minipage}[b]{0.13\linewidth}
  \centering
  \centerline{\includegraphics[width=1\linewidth,height=0.9\linewidth]{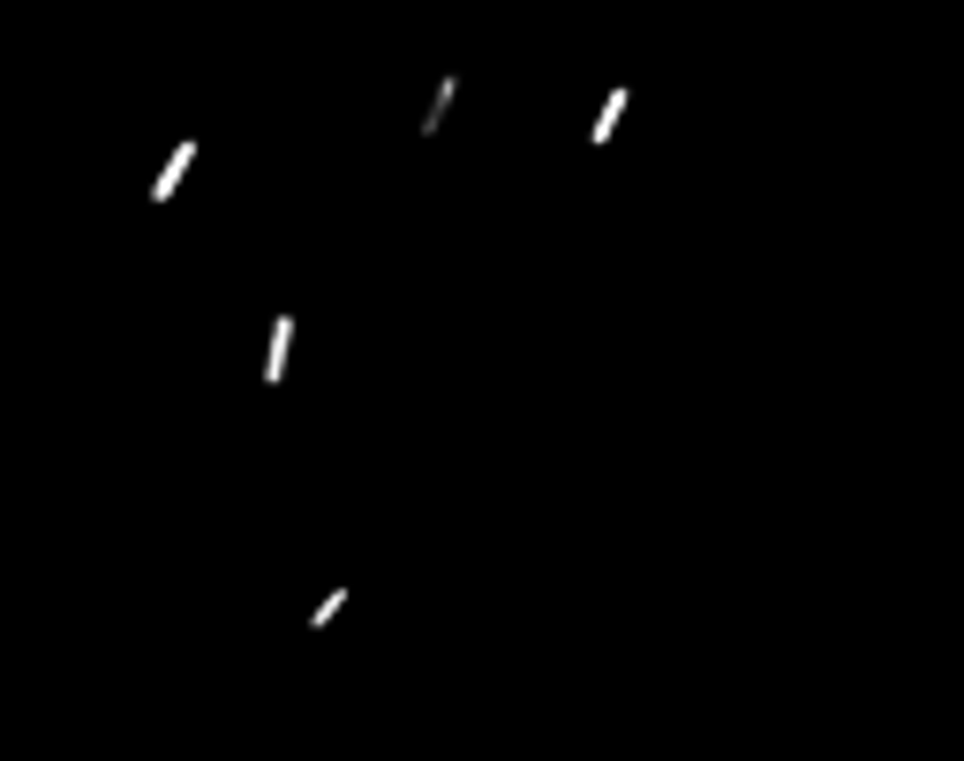}}
\end{minipage} \\

\begin{minipage}[b]{0.13\linewidth}
  \centering
  \centerline{\includegraphics[width=1\linewidth,height=0.9\linewidth]{0194.jpg}}
\end{minipage}
\begin{minipage}[b]{0.13\linewidth}
  \centering
  \centerline{\includegraphics[width=1\linewidth,height=0.9\linewidth]{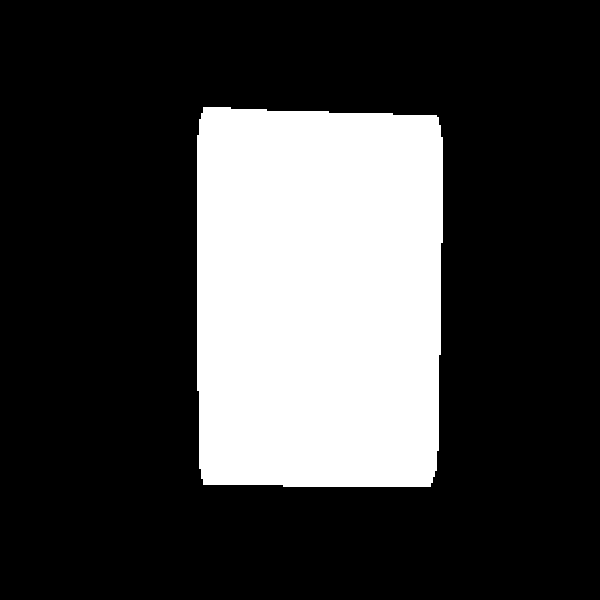}}
\end{minipage}
 \begin{minipage}[b]{0.13\linewidth}
  \centering
  \centerline{\includegraphics[width=1\linewidth,height=0.9\linewidth]{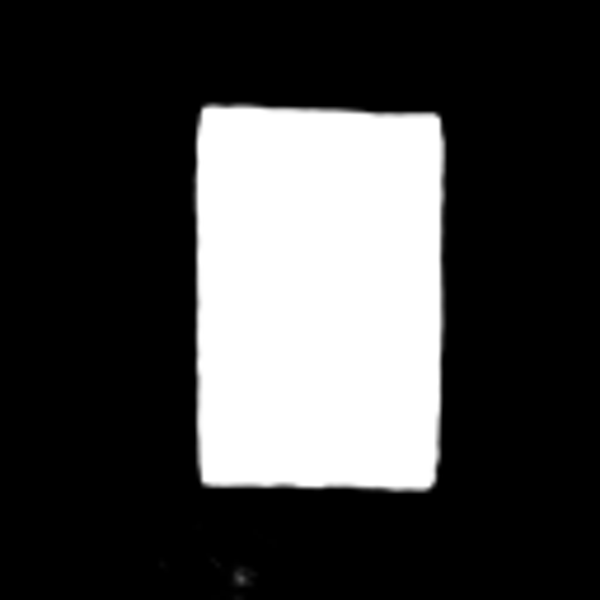}}
\end{minipage}
 \begin{minipage}[b]{0.13\linewidth}
  \centering
  \centerline{\includegraphics[width=1\linewidth,height=0.9\linewidth]{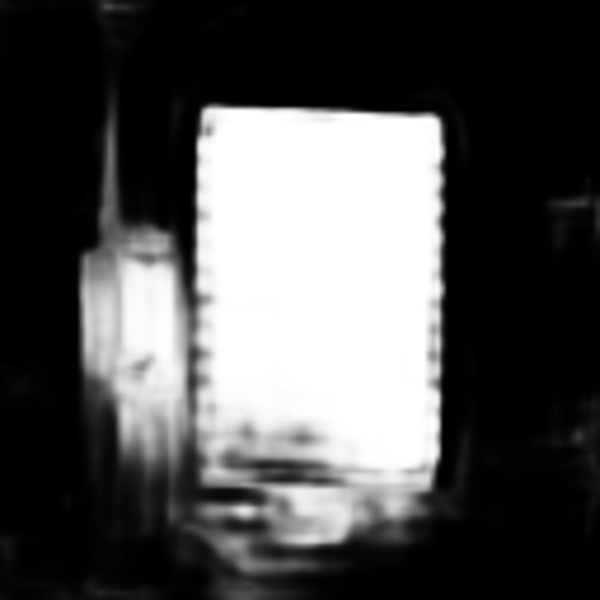}}
\end{minipage}
 \begin{minipage}[b]{0.13\linewidth}
  \centering
  \centerline{\includegraphics[width=1\linewidth,height=0.9\linewidth]{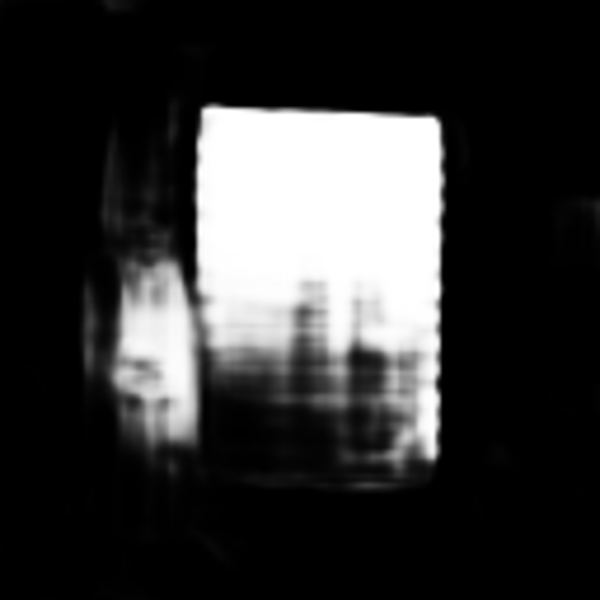}}
\end{minipage}
 \begin{minipage}[b]{0.13\linewidth}
  \centering
  \centerline{\includegraphics[width=1\linewidth,height=0.9\linewidth]{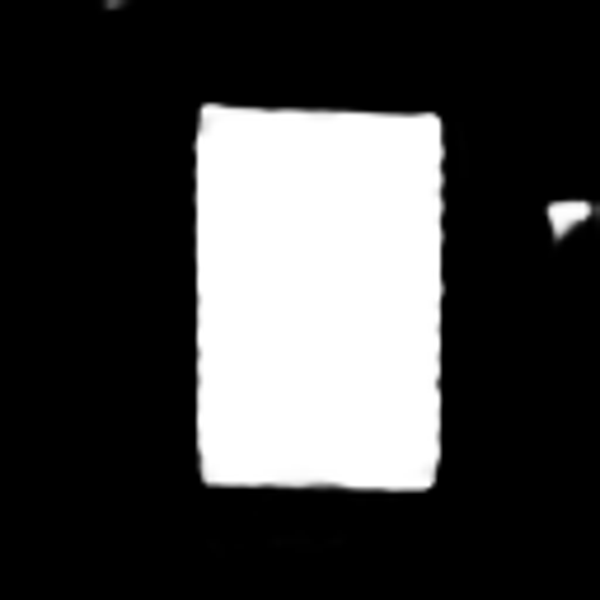}}
\end{minipage}
 \begin{minipage}[b]{0.13\linewidth}
  \centering
  \centerline{\includegraphics[width=1\linewidth,height=0.9\linewidth]{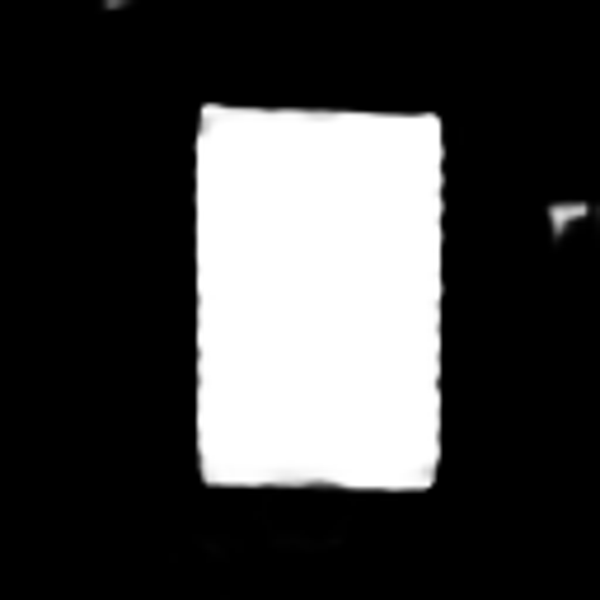}}
\end{minipage} \\
\caption{Visual results of ablation study. From left to right are input optical RSIs, the corresponding ground truth, the results of the proposed PDF-Net, PDF-Net w/o DC, PDF-Net w 1-DC, PDF-Net w/o CPC, and PDF-Net w/o DUS.}
\label{fig:ablaiton}
\end{figure}

In Table \ref{tab2}, we can see that our proposed PDF-Net achieves the best quantitative results than the modified versions, which further demonstrates the effectiveness of each key component used in our network. In addition, the PDF-Net w/o DC obtains the worst performance, which indicates the importance of the dense connections for SOD in optical RSIs. The main reason is that the dense connections can combine the detail information with the semantic information. Both of them are significant for SOD in optical RSIs. Besides, we also provide visual comparison results in Fig.~\ref{fig:ablaiton}.

As shown in Fig.~\ref{fig:ablaiton}, the salient objects in the saliency maps of our PDF-Net have the complete  structure and clear boundary. Additionally, our PDF-Net can predict the salient objects, even the objects with large or small scales. In comparison, the PDF-Net w/o DC and PDF-Net w 1-DC cannot suppress the cluttered backgrounds and thus leave the non-salient objects in the results. Moreover, these two networks cannot produce the complete structure of salient objects, which further indicates the importance of dense connections. In comparison to the final PDF-Net results, the PDF-Net w/o CPC and PDF-Net w/o DUS cannot perceive small-scaled objects that are obvious in the results of the second line. Moreover, the results of them look not clear.

To sum up, the ablation studies indicate that a) the dense connections are useful for SOD in optical RSIs, and b) the down-up sampling operations can perceive scaled objects well; and c) more reasonable network design is needed, especially for the challenging SOD in optical RSIs.

\vspace{5 mm}
\section{Conclusion}\label{sec6}
We propose a specially designed SOD network for optical RSIs, called PDF-Net, in this paper, which can effectively and accurately detect the diversely scaled salient objects in optical RSIs by making use of the in-path and cross-path information and the multi-resolution features. The proposed PDF-Net consistently outperforms state-of-the-art SOD methods on the ORSSD dataset in terms of visual comparisons and quantitative evaluations. Besides, the effectiveness of key components is verified in the ablation analysis, which further demonstrates the rationality and effectiveness of the proposed network.

\par

\bibliography{mybibfile}

\end{document}